\documentclass[conference]{IEEEtran}
\IEEEoverridecommandlockouts
\usepackage{cite}
\usepackage{amsmath,amssymb,amsfonts}
\usepackage{algorithmic}
\usepackage{graphicx}
\usepackage{textcomp}
\usepackage{xcolor}

\usepackage{hyperref}
\usepackage{subcaption}
\usepackage{enumitem}
\usepackage[font=small]{caption}
\usepackage{balance}
\usepackage{booktabs}

\def\BibTeX{{\rm B\kern-.05em{\sc i\kern-.025em b}\kern-.08em
    T\kern-.1667em\lower.7ex\hbox{E}\kern-.125emX}}

\newcommand{\tocite}[1]{\textcolor{red}{[CITE]}}
\newcommand{\todo}[1]{\textcolor{orange}{[TODO]}}
\newcommand{\tocheck}[1]{\textcolor{cyan}{[CHECK]}}

\begin{document}


\title{OfficeMate: Pilot Evaluation of an Office Assistant Robot
\thanks{This research is partially supported by the Australian Research Council Discovery Early Career Research Award (Grant No. DE210100858).}
}




\author{
\IEEEauthorblockN{Jiahe Pan}
\IEEEauthorblockA{
\textit{The University of Melbourne}\\
Melbourne, Australia \\
michael.pan@unimelb.edu.au}
\and
\IEEEauthorblockN{Sarah Sch\"{o}mbs}
\IEEEauthorblockA{
\textit{The University of Melbourne}\\
Melbourne, Australia \\
sschombs@student.unimelb.edu.au}
\and
\IEEEauthorblockN{Yan Zhang}
\IEEEauthorblockA{
\textit{The University of Melbourne}\\
Melbourne, Australia \\
yan.zhang.1@unimelb.edu.au}
\and[\hfill\mbox{}\par\mbox{}\hfill]
\IEEEauthorblockN{Ramtin Tabatabaei}
\IEEEauthorblockA{
\textit{The University of Melbourne}\\
Melbourne, Australia \\
stabatabaeim@student.unimelb.edu.au}
\and
\IEEEauthorblockN{Muhammad Bilal}
\IEEEauthorblockA{
\textit{The University of Melbourne}\\
Melbourne, Australia \\
muhammad.bilal1@student.unimelb.edu.au}
\and
\IEEEauthorblockN{Wafa Johal}
\IEEEauthorblockA{
\textit{The University of Melbourne}\\
Melbourne, Australia \\
wafa.johal@unimelb.edu.au}
}

\maketitle

\begin{abstract}

Office Assistant Robots (OARs) offer a promising solution to proactively provide in-situ support to enhance employee well-being and productivity in office spaces. We introduce OfficeMate, a social OAR designed to assist with practical tasks, foster social interaction, and promote health and well-being. Through a pilot evaluation with seven participants in an office environment, we found that users see potential in OARs for reducing stress and promoting healthy habits and value the robot’s ability to provide companionship and physical activity reminders in the office space. However, concerns regarding privacy, communication, and the robot’s interaction timing were also raised. The feedback highlights the need to carefully consider the robot’s appearance and behaviour to ensure it enhances user experience and aligns with office social norms. We believe these insights will better inform the development of adaptive, intelligent OAR systems for future office space integration.

\end{abstract}


\section{Introduction}



In today's fast-paced work environment people spend the majority of their week in the office -- sitting at a desk, attending meetings, and interacting with others in the office. To support their work life, health and well-being, companies increasingly provide services and activities such as gym memberships or team events. While helpful, these perks can be inaccessible and time-consuming, and may require active behavioural adaptations from employees to spend additional time outside of work.

Office assistant robots (OAR) offer immense potential to provide such services in situ, to reduce stress at work and to enhance the overall well-being of employees. As conversation partners, OARs could offer words of encouragement, jokes and assistance as well as foster a collaborative work environment. By promoting positive social interactions, these robots could help cultivate empathy and pro-social behaviour \cite{schombs2023feeling} among employees and enhance interpersonal relationships in the office. They could further encourage healthy habits by providing nutritious snacks and water \cite{gouko:2017}, communicating health risks related to physical inactivity \cite{schombs:2024}, or leading mindfulness exercises \cite{batuhan:2024,spitale:2023} during breaks. Beyond being social, OARs could also provide support for every-day tasks and take over repetitive responsibilities that are time-consuming and interruptive to workflows (e.g. directing clients to meeting rooms, keeping track of meetings), allowing employees to prioritise tasks that are relevant to their domain. However, developing such an adaptive, intelligent, and holistic system that is suitable for the office environment is challenging. 

A variety of studies have explored the development of OARs, each focusing on unique applications to enhance workplace efficiency and user experience. 
For instance, \cite{8736141} describes an office robot capable of autonomously navigating and delivering documents and small parcels to employees, who can use an Android app for voice or virtual command control of the robot. \cite{10.1007/978-3-030-97672-9_17} presents a contactless OAR aimed at reducing face-to-face interactions, particularly relevant during the COVID-19 pandemic. This robot acts as a digital receptionist, using facial recognition to personalise interactions, provide directions, and answer common inquiries while minimising physical contact.

Given the potential of robots to engage office workers and to provide health and well-being support, we developed a functional OAR system called OfficeMate that allows to explore the design considerations of an effective and socially-acceptable OAR. Through a pilot user evaluation in a real office environment, we aim to gain insights into how people may perceive the existence of an OAR in the office, including both benefits and drawbacks. These insights could potentially better inform the design of OARs to allow for a smoother integration of such systems into real office spaces.
Overall, we aim to contribute in the following aspects: 1) integrate key functional components into an effective OAR system (and open-source our implementations), 2) conduct a preliminary evaluation of the system within a real office environment, 3) discuss our findings and provide insights on how to improve the system for future office space integration.
\section{The OfficeMate: Functions and Implementation}



The OfficeMate is based on the TIAGo mobile robot, which is well-supported with ROS libraries including Navigation \cite{guimaraes2016nav}, MoveIt \cite{chitta2012moveit}, and a text-to-speech (TTS) interface. ROS was used for communication between the system components. In this section, we describe our implementation of each module of OfficeMate and describe their integration\footnote{The source code and a link to a demo video are available in the project repository on GitHub: \href{https://github.com/yzhang2332/woa\_tiago}{https://github.com/yzhang2332/woa\_tiago}.}.

\begin{figure}
    \centering
    \includegraphics[width=\linewidth]{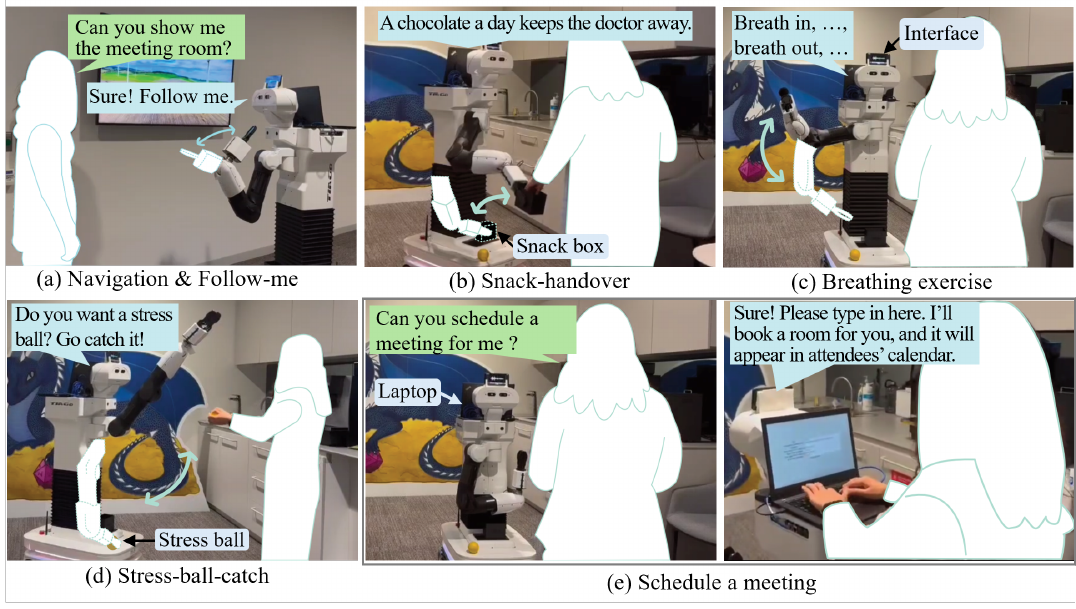}
    \caption{Examples of interactions with OfficeMate's functionalities.}
    \label{fig:interaction}
\end{figure}


\subsubsection{Natural Conversation}\label{subsubsec:natural-conversation}
As an OAR, engaging in friendly and empathetic social interactions with humans is essential. A key component of social interactions is natural language conversation, where the OAR perceives and understands the user’s speech, determines necessary actions, and responds naturally. We implemented natural verbal communication by integrating the Large-Language Model (LLM), GPT4, which has been used in HRI contexts previously \cite{zhang2023large}, and the TTS interface of the TIAGo robot. The user's speech, captured via a ReSpeaker USB Microphone Array, is converted to text with GPT-4's speech-to-text, then checked against a database of trigger words which map to a predefined set of actions. If a match is found, the action executes, and GPT-4 generates a natural language response to the user, which is played through the robot's TTS during or after the action.

To facilitate a more engaging interaction and effective communication, we designed and incorporated a proactive behaviour where the robot suggests a random subset of its functions if no trigger words are detected after a set time period. Furthermore, to help users understand whether the robot is listening or speaking, we created a browser-based interface which uses two GIFs to show whether the robot is listening or speaking (Fig. \ref{fig:interaction}c). The interface is displayed on a phone which is mounted on top of TIAGo's head. 

Overall, this module acts as the ``brain" of the robot, and is responsible for both reasoning about the required tasks and actions through natural language and maintaining a friendly and engaging interaction with the user.

\subsubsection{Person-Following Behaviour} \label{subsubsec:person-following}
Existing works in HRI have shown that a mobile service robot exhibiting a person-following behaviour can lead to a more natural interaction and socially-acceptable perception from the user \cite{honig2018toward}. 
We use YOLOv8 \cite{yolov8_ultralytics} to detect the target person and estimate their location relative to the robot. Then, we apply a direct-following \cite{chen2007person} approach, using constant velocity \cite{koide2016identification} commands to the robot's base to keep the target person centred horizontally within the camera frame and to maintain a constant distance from them. Threshold values are used to enhance robustness against ambiguous person detections and noisy depth data. A simple reactive collision avoidance algorithm using the robot's laser scans of its local environment is also incorporated. The person-following behaviour is always active except when another action which controls the robot's base movements is also active, in which case the latter takes priority.

\subsubsection{Custom Gestures and Activities} \label{subsubsec:gestures}
Co-verbal gestures often help us convey ideas and facilitate engaging conversations. They have also been shown to be effective in attracting and maintaining listeners' attentions \cite{bremner2011effects} in HRI contexts. For OfficeMate, we designed a custom gesture library which can be customised and extended based on the interaction scenario and robot design. Specifically, there are three co-verbal gestures: welcome, follow-me (Fig.\ref{fig:interaction}a), and show-around, which are associated with the behaviours of welcoming a visitor, navigating them to a desired location, and showing them around once arrived, respectively. Moreover, prior works have shown that activities such as deep-breathing \cite{tavoian2023deep} and meditation \cite{pagliaro2020effects}, or having food-breaks \cite{corvo2020eating}, can positively impact employees' well-being in an office environment. Thus, OfficeMate is equipped with three activity-based gestures: snack-handover (Fig.\ref{fig:interaction}b), breathing exercise (Fig.\ref{fig:interaction}c), and stress-ball-catch (Fig.\ref{fig:interaction}d). All activities can be evoked by the user through both explicit and implicit language instructions. 

\subsubsection{Autonomous Navigation}\label{subsubsec:auto-navigation}
Another key functionality of an OAR is navigating visitors to their desired locations (e.g. meeting room, coffee machine) within the office space. Our implementation used the built-in mapping, localization, and path planning functionalities of the TIAGo robot to generate collision-free trajectories to any feasible navigation goal. Users can initiate the navigation functionality directly through natural language instructions which contain a single location keyword (e.g. meeting room, kitchen). In addition, we provide a screen-based interface which allows users to record and associate a location keyword with the current robot pose, which they can later query as a navigation goal.

\subsubsection{Performing Office Tasks}\label{subsubsec:office-task}
As an OAR, the OfficeMate must also effectively perform office tasks. These may include sending emails, scheduling meetings, and providing calendar reminders. We used the Google API and created a screen-based GUI using Python which allows the user to create meetings in their own calendar and invite other attendees through email. Since the laptop is mounted on the TIAGo's back, the robot executes a half-turn to face the screen towards the user, before turning back once they have scheduled the meeting (Fig.\ref{fig:interaction}e). Furthermore, OfficeMate is also capable of providing reminders of scheduled meetings in the user's calendar upon request. 

\section{Pilot Evaluation}

We conducted a series of in-person pilot evaluations with seven participants in an office environment. The primary goals of the pilot were to gain insight into how participants envision integrating such a robot into their office space, to determine whether they could envision such a system being deployed in the office to support employee well-being, and to identify any concerns, reservations, and areas for improvement.
The pilot studies were approved by the University's ethics committee.

\subsection{Task Setup and Recruitment}

The robot was deployed in the university's open office space. We recruited participants who were both students and staff members. In each session, participants were first introduced to the study and asked to sign a consent form. They were then introduced to the OfficeMate, with an overview of its functionalities provided through a user manual. Next, participants received basic information relating to the interaction, including saying ``Hey TIAGo" to initiate interactions, and observing the mobile phone screen to confirm when OfficeMate was actively listening or responding. This was followed by the main interaction phase, where participants were encouraged to engage with all of OfficeMate's functionalities, i.e. natural conversation, navigation guidance, mindfulness and physical exercises, and snack and coffee machine assistance.. Finally, each session concluded with an approximately 10 min, semi-structured interview to gather insights on the participants' experience. 



\subsection{Qualitative Results}
The semi-structured interview began with ice-breaking questions about participants' prior experience with assistive robots and general interest in robots in an office context, followed by several in-depth questions on their perceptions and preferences regarding the robot's role in supporting mental and physical health in the office after their interaction with it. Lastly, participants were asked to express their concerns and elaborate on potential challenges they identify for the deployment of such assistant robots in an office environment.

To analyse the qualitative data, we transcribed and revised the recorded audio files and familiarised ourselves with the data by reading the transcripts. We identified relevant quotes and visually organised them on a Miro board using affinity mapping, a common method in HCI to cluster qualitative data and derive insights. Guided by our research objectives, we used these as tentative themes to form an initial structure \cite{terry_data_2021}; however, the analysis was predominantly inductive due to the exploratory nature of this pilot study \cite{braun_doing_2022}. We remained open to novel insights beyond the predefined objectives. From the identified quotes, we formulated codes and iteratively refined both the clusters and thematic categories to ensure an alignment with our evolving understanding of the data.


\subsection{Discussion of Results}

\subsubsection{Support for Mental and Physical Well-Being}\label{discuss1}
The majority of participants felt that having an OAR in the office space can support employees' well-being, both mental and physical.

``I think that as most office workers sit in chairs for hours, the robot can help overcome this challenge. They can come to employees, talk to them, and encourage physical activities.'' -- P6

The mindfulness exercise and the robot's ability to offer snacks and guide visitors around the office were perceived as particularly stress-reducing by the participants; e.g. ``Yeah, I think it's helpful to play the relaxing exercise with the robot.'' -- P2; 
They further emphasised the robot's role \textbf{to act as a well-being reminder}, whether it's an exercise or to take a break. However, two participants expressed the need for the robot to \textbf{initiate well-being practices at the right time}, highlighting that the robot should a) play a proactive role in the office well-being management and b) do so at the right time, without distracting them. Interestingly, one participant highlighted that the robot is increasing their motivation to care for their well-being by \textbf{actively taking part in the exercise as well} (``It's like you have someone to do the relaxing exercise with you, so I can be more motivated ... it's more fun, you have some socialisation.'' -- P2). This companion effect was also highlighted by P2, who described how the robot's action of coming to their office door and inviting them to take a break and walk together would feel supportive and reassuring. 

``[The robot] coming to your office door and saying, "Let's have a little break, walk with me." Just walking in the corridor together. It doesn't have to say anything, but just having someone come to you and offer you to do something''. 

Whereas several participants expressed the robot's potential to enhance their productivity by supporting their mental health or taking over interruptive tasks such as guiding visitors to the meeting room, one participant expressed a potential trade-off between productivity and physical activity. While the robot allows them to continue working without interruption, it may reduce opportunities for physical movement, as they would otherwise have walked to the meeting room themselves. This reflects a possible unintended consequence: increased productivity at the expense of reduced physical activity.


\subsubsection{Concerns Around the Deployment of Office Assistant Robots in the Office}\label{discuss2}

Some participants expressed concerns around the robot's noise possibly disturbing the office, ``whole-body movement'' (P1) or inadequate behaviour such as telling jokes at inappropriate moments (P2), which shows the need for the OAR to not just identify the appropriate timing to initiate interaction, but to align with office norms (e.g. ``giving it a personality that is adequate''- P2). The deployment of an OAR in the office further raised several concerns around privacy (P1,P3,P4); e.g. ``I think people might end up feeling like they're being monitored. What is the robot learning about me? Is this information coming back to my boss?''. P3 further questioned the authenticity of the robot's well-being support, noting that if it is indeed imposed or managed externally (by HR), it could feel intrusive or insincere, similar to mandatory social activities. Contrary to the aforementioned benefit of companionship, P6 expressed the concern that an increased reliance on an OAR for workplace interactions might reduce direct communication between employees and potentially impact interpersonal relationships among staff.

\subsubsection{Concerns Around the Robot System}\label{discuss3}
Participants also highlighted specific aspects of the robot itself that could be improved to enhance usability and comfort. Some found it challenging to anticipate the robot’s movements (robot arm in particular) (P2, P3, P4, P5), which raised safety concerns, while two participants noted that the robot’s movements were too slow (P4), which affected the interaction flow. Two participants expressed frustration with the conversation features (P2, P5), critiquing latency issues, slow response times, and the need to speak loudly. In addition, one participant suggested that a more soothing voice would be better suited for guiding mindfulness practices (P7), and therefore the need to align the robot’s tone with its well-being role.

Interestingly, participants shared mixed opinions regarding the robot's appearance, especially its suitability for an office environment. One participant suggested that a pet-like design, such as a robotic dog or cat, is appealing and might be more engaging for well-being support; ''I think a robotic dog might have a better chance at developing well-being.'' -- P1. Contrary, one participant suggested to include more gestures and expressions to the robot design to make it even more human-like (P3). Others felt that the current design proposes a good degree of humanlikeness (P4, P7) and appreciated the practicality of the robot's arm, which enables it to assist more effectively by handing over items, while also highlighting the trade-off between functionality and appearance (P5).

\section{Recommendations and Future Directions}

We aim for our implementation of the functionalities and behaviours of OfficeMate to serve as a baseline of an OAR system for future works to build upon. Based on our results, we recommend two specific aspects for improvement, and provide an outlook on integrating OARs which provide practical office and well-being assistance into real office environments.

\subsection{Improving Communication}
The results indicate that the communication between the human and the OAR could be improved by increasing the efficiency and robustness of the speech detection and generation modules which could lead to more fluid verbal interactions. Other possible enhancements include equipping the OAR with a more natural and adaptive voice which suits the current context, and adjusting the robot to a suitable height to facilitate comfortable communication with a human user \cite{hiroi2016influence}. 

Moreover, the safety concern around fast robot movements expressed by some participants shows the importance for an OAR to clearly communicate its intent to the human prior to acting. Prior research has explored using expressive robot behaviour to indicate motion and navigation intent, including light signals, gaze, head pan or gripper movements \cite{lemasurier:2021,fernandez:2018,may:2015}, which can be considered in future iterations to increase interpretability and safety in office spaces. However, it is worth highlighting that other participants instead preferred faster robot arm movements for more efficient interaction. This therefore suggests the potential need for the OAR to have adaptive behaviours which are personalised for different users' preferences. Here, recent works have shown LLMs' ability to generate expressive and adaptive robot motions through learning from human feedback during the interaction \cite{mahadevan2024generative}. 

\subsection{Adhering to Social and Office Norms}
The participants' feedback underlines the importance for an OAR to adhere to social and office space norms. First, the OAR should not be distracting towards the employees by identifying the right timing to propose well-being activities, and to do so in a natural way such that employees feel encouraged and motivated instead of forced or obligated. Results also show the importance for the OAR to adhere to social norms by ensuring that it remains attentive and approachable while not appearing to be fixating on employees, where its presence in the office may raise privacy concerns. This may also help ensure that the perceived role of the OAR by employees and its actual role within the social context of the organisation are aligned \cite{10.1145/3613905.3651002}. 
Furthermore, feedback from the study resonates with previous findings that both a human-like appearance and facial expressions can enhance user experience \cite{ringwald2023should}. Therefore, it is worth investigating the effectiveness of different robot morphologies and appearances in enhancing the social acceptance of an OAR in the office, without negatively impacting its functionality of providing assistance in office tasks.

Moreover, in a dynamic office environment with potentially crowded spaces, additional socially-motivated objectives for navigation may arise \cite{sehestedt2010robot}. These include generating socially-acceptable approach trajectories to humans, tracking the position of the target human, and leading them to their destination in an appropriate manner \cite{chebotareva2020basic}. Here, socially-aware robot navigation have also been shown to increase adherence to the same norms of the pedestrians around the robot \cite{johnson2018socially}.

\subsection{Providing Practical Assistance and Future Directions}
Despite the rapid advancements in both robot hardware and the control and learning algorithms, current robots are still far from achieving the same level of dexterity and versatility as humans in complex manipulation and locomotion tasks \cite{riener2023robots}, especially in confined or crowded spaces such as a typical office environment. 
Therefore, successfully deploying a holistic robotic system that is able to provide practical assistance in a real office environment is challenging and still requires further scientific progress in many sub-fields of robotics. Nevertheless, through presenting an in-situ evaluation of OfficeMate, we aim to highlight the importance of taking into account the human user when innovating in each of these aspects of a robotic system. 
Our pilot study with 7 participants offers valuable first insights into how users in an office space may interact with an office assistant robot, including feedback on the OfficeMate's components, user concerns, and areas for improvement. Ultimately, we hope that these findings will guide the next phase of implementation and design, incorporating user feedback into an iterative approach to refining and improving the OfficeMate for real-world deployment.
\bibliographystyle{ieeetr}
\balance
\bibliography{officemate}

\end{document}